\documentclass[runningheads]{llncs}
\usepackage[T1]{fontenc}
\usepackage{graphicx}
\usepackage{subcaption}
\usepackage{amsmath}
\usepackage{booktabs}
\usepackage{multirow}
\usepackage{array}
\usepackage{enumitem}
\usepackage{url}
\usepackage{amsmath}   
\usepackage{amssymb}   
\usepackage{amsfonts} 
\usepackage{mathtools} 
\newcommand{\eg}{e.g.}

\usepackage{xcolor, colortbl}
\usepackage{cite}
\usepackage{hyperref}

\begin{document}
\title{Towards Label-Free Single-Cell Phenotyping Using Multi-Task Learning}
\author{Saqib Nazir \and
Ardhendu Behera}
\authorrunning{S. Nazir et al.}
\institute{Department of Computer Science, Edge Hill University, UK \\
\email{(Saqib.Nazir,Ardhendu.Behera)@edgehill.ac.uk}}
\maketitle
\begin{abstract}
Label-free single-cell imaging offers a scalable, non-invasive alternative to fluorescence-based cytometry, yet inferring molecular phenotypes directly from bright-field morphology remains challenging. We present a unified Deep Learning (DL) framework that jointly performs White Blood Cell (WBC) classification and continuous protein-expression regression from label-free Differential Phase Contrast (DPC) images. Our model employs a Hybrid architecture that fuses convolutional fine-grained texture features with transformer-based global representations through a learnable cross-branch gating module, enabling robust morpho-molecular inference from DPC images. To support downstream interpretability, we further incorporate a Large Language Model (LLM) that generates concise, biologically grounded summaries of the predicted cell states.
Experiments on the Berkeley Single Cell Computational Microscopy (BSCCM) and Blood Cells Image benchmarks demonstrate strong performance, achieving a 91.3\% WBC classification accuracy and a 0.72 Pearson correlation for CD16 expression regression on BSCCM. These results underscore the promise of label-free single-cell imaging for cost-effective hematological profiling, enabling simultaneous phenotype identification and quantitative biomarker estimation without fluorescent staining. The source code is available at 
\href{https://github.com/saqibnaziir/Single-Cell-Phenotyping}{https://github.com/saqibnaziir/Single-Cell-Phenotyping}.
\end{abstract}

\section{Introduction}
Accurate characterization of White Blood Cells (WBCs) is essential for hematologic diagnosis, as alterations in cell morphology and protein-expression profiles are key indicators of immune dysregulation, infection, and hematologic malignancies. Conventional workflows rely on fluorescence-based flow cytometry or manually stained smear examination, both of which have practical limitations: flow cytometry requires specialized instruments and fluorophore labeling, while manual microscopy is labor-intensive, subjective, and inherently low throughput\cite{tomkinson2024toward}.

Recent studies demonstrate that label-free optical modalities, particularly Quantitative Phase Imaging (QPI) and Differential Phase Contrast (DPC), encode detailed biophysical information correlated with cellular identity and functional state. QPI has allowed label-free WBC classification in holographic flow cytometry \cite{Ciaparrone2024} and hematological profiling with optical diffraction tomography \cite{Ryu2023}. Deep Learning (DL) applied to these modalities has shown that subtle high-frequency morphological variations can be predictive of cell subtype and activation state \cite{Rivenson2019,Li2019DPC, nazir2026hybrid}. These findings position label-free imaging as a promising, low-cost alternative to staining-based cytometry. However, most of the existing techniques are limited in two important aspects.
First, prior work predominantly treats WBC analysis as a discrete classification problem, overlooking the continuous spectrum of protein-expression levels that quantify molecular function. Direct regression of surface-marker intensities (e.g., CD16, CD45) from label-free imagery remains largely unexplored, despite initial evidence from imaging flow cytometry and Raman-based profiling that morphology can correlate with underlying molecular signatures \cite{zhang2022ifcprotein,Raman2RNA2024}.
Second, the black-box nature of deep neural networks restricts clinical adoption. Models often produce predictions without interpretable biological reasoning, and clinical readiness requires transparent mechanisms that support expert validation and explainability.

In this work, we address the broader challenge of \textit{morpho-molecular inference} predicting both discrete cell type ($y_{\text{cls}}$) and continuous protein-expression levels ($y_{\text{reg}}$) directly from a single-cell label-free DPC image. This is a fundamentally difficult problem, as protein associated morphological cues are subtle, sparsely distributed, and easily attenuated by standard convolutional pooling.
To overcome these challenges, we propose a unified multi-task learning framework that captures complementary morphological cues. Our key novelty lies in three architectural innovations: First, we introduce a \textit{dual-branch hybrid encoder} that combines CNN-based local texture extraction with transformer-based global features, addressing the limitation that purely convolutional architectures fail to capture long-range morphological patterns critical for protein-expression inference. Second, we design a task-adaptive \textit{gating mechanism} that dynamically modulates feature sharing between classification and regression pathways, effectively mitigating negative transfer that typically degrades multi-task performance when tasks have conflicting gradient dynamics. Third, we integrate an LLM-based interpretation module that translates model predictions into biologically grounded explanations, advancing model transparency beyond conventional attention visualization.
To our knowledge, this is one of the first works to jointly regress marker-level protein expression and classify WBC phenotype from label-free DPC microscopy, supported by per-marker evaluation and LLM-generated biological reasoning.
Our main contributions are:
\vspace{-3mm}
\begin{enumerate}
\item A unified multi-task framework that jointly predicts WBC class and protein-expression levels from label-free DPC images, eliminating the need for chemical staining.
\item A dual-branch hybrid encoder combining multi-scale convolutional texture features with transformer-based global representations via a learnable fusion layer.
\item A task-adaptive gating mechanism that balances shared and task-specific information, improving both classification and regression accuracy.
\item Comprehensive experiments on the BSCCM and Blood Cells Image dataset (BCCD) demonstrating improvements over single-task and purely convolutional baselines.
\end{enumerate}
\vspace{-6mm}
\section{Related Work}
\vspace{-3mm}
\subsection{Label-free and Stained WBC Classification}
\vspace{-2mm}
DL has shown strong performance in stained WBC classification, with VGG variants \cite{simonyan2015very} and custom CNNs achieving high accuracy on datasets such as BCCD and Raabin-WBC \cite{raabin2021large}. However, staining introduces chemical variability and precludes live-cell imaging. Label-free modalities such as DPC and QPI preserve native cell morphology but are more challenging due to lower contrast. Early approaches used handcrafted textural descriptors \cite{habibzadeh2021review}, while recent work applied DL to holographic QPI sorting and feature fusion for DPC microscopy \cite{Ryu2023}. However, these efforts primarily address classification, while our work explores a label-free morphology that also encodes a sufficient signal for continuous protein-expression regression, a significantly difficult and understudied task.
\vspace{-3mm}
\subsection{Protein-Expression Prediction and Multi-Modal Analysis}
Virtual staining methods demonstrate that label-free imaging can recover fluorescence-like contrast \cite{Rivenson2019}, and pathological studies show that morphology can weakly encode molecular states such as mutations \cite{naylor2018segmentation}. Beyond image translation, recent works predict molecular signatures from other modalities: imaging flow cytometry models infer surface-marker levels from fluorescence channels \cite{zhang2022ifcprotein}, while Raman-based models regress transcriptomic profiles \cite{Raman2RNA2024}. These are the closest in spirit to our goal, but they rely on multi-channel fluorescence or spectral input, not single-cell label-free DPC images. In contrast, we perform direct regression of protein-expression levels from a single bright-field modality, jointly with cell-type classification. Architecturally, previous hybrid CNN–Transformer models such as TransUNet \cite{chen2021transunet} and MedT \cite{valanarasu2021med} improve morphological feature extraction; we build on this line by designing a multi-task hybrid encoder with task-adaptive gating for morpho-molecular inference.
\vspace{-3mm}
\subsection{Language Models for Biomedical Summarization}
LLMs have been used for structured radiology summarization \cite{yan2023style} and biomedical text generation \cite{luo2022biogpt,li2023clinicalt5}. However, their integration into single-cell microscopy pipelines remains unexplored. Our approach employs an LLM to convert quantitative predictions into concise biological descriptions. Given the known risks of hallucination in LLMs, we constrain the module via template-based prompting and restrict output to grounded summaries derived solely from model predictions.
\vspace{-5mm}
\begin{figure}[t]
\centering
\includegraphics[width=\linewidth]{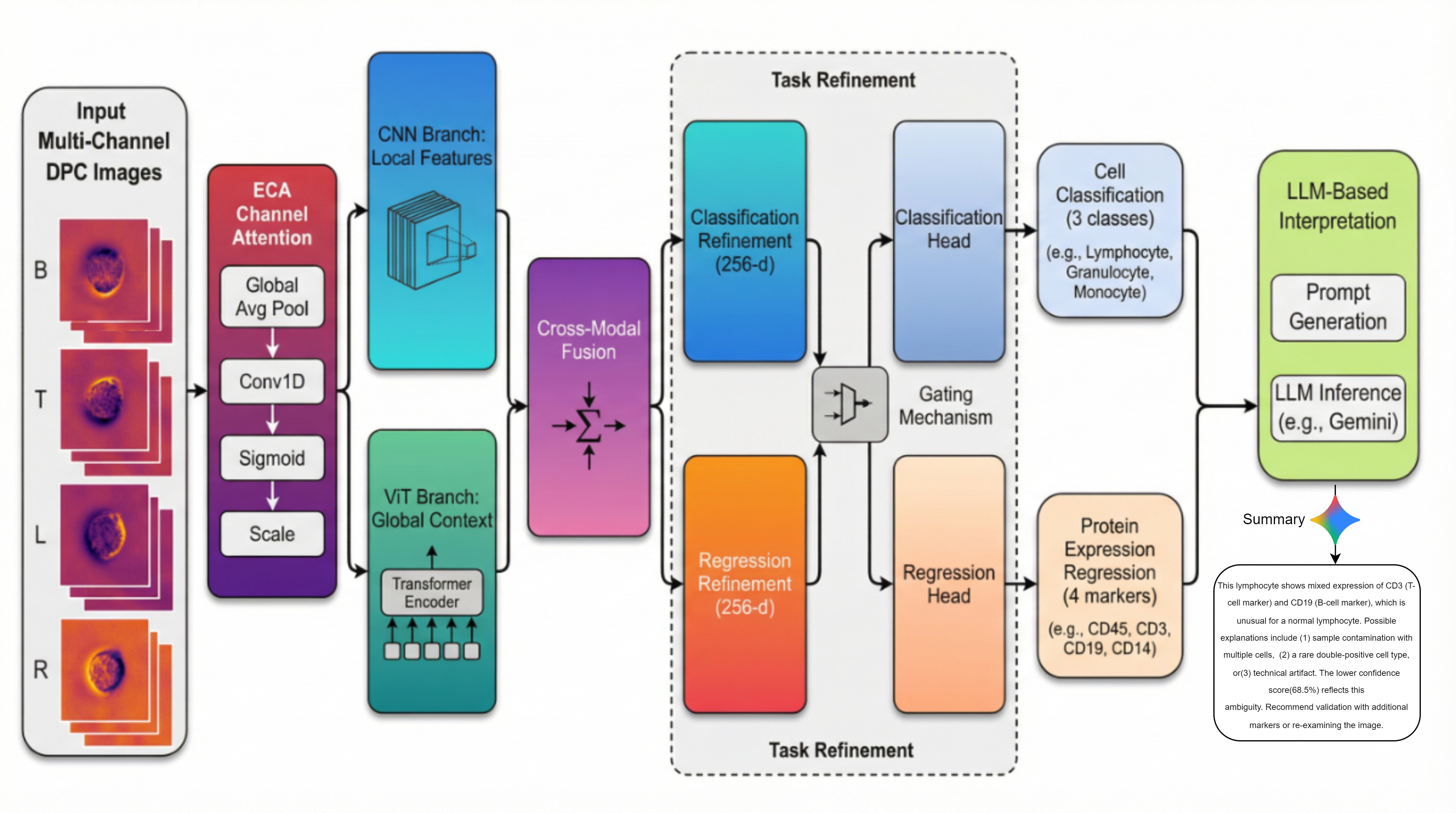}
\caption{The proposed hybrid CNN-ViT architecture: a CNN extracts local spatial features from label-free DPC images, a ViT module captures global dependencies via self-attention, and dual task-specific heads perform WBC classification and continuous protein-expression regression. Feature fusion between CNN and ViT branches enables complementary local-global representation learning.}
\label{fig:model}
\vspace{-5mm}
\end{figure}

\section{Proposed Method} 
\label{sec:method} 
\vspace{-2mm}
Given a multi-channel DPC image $\mathbf{X} \in \mathbb{R}^{B \times 4 \times 28 \times 28}$ where $B$ is the batch size and the four channels correspond to left, right, top, and bottom illumination directions (input shown in Fig \ref{fig:model}), our objective is to simultaneously predict: (1) cell type labels $\mathbf{y}_{\text{cls}} \in \{0,1,2\}$ representing WBC, (2) protein expression levels $\mathbf{y}_{\text{reg}} \in \mathbb{R}^4$ for markers CD45, CD3, CD19, and CD14. This dual-task formulation enables the model to take advantage of shared representations while maintaining task-specific specialization. 
% \subsection{Efficient Channel Attention (ECA)} 
DPC microscopy produces multi-directional illumination channels that contain redundant and complementary information. We employ Efficient Channel Attention (ECA)~\cite{wang2020eca} to adaptively weight the four DPC illumination channels. 

\noindent\textbf{CNN Branch:}
The CNN branch is designed to capture fine-grained morphological cues that dominate single-cell DPC imaging. We use a $3{\times}3$ stem (4$\rightarrow$64) followed by Inception-style modules with residual connections~\cite{szegedy2015rethinking}. Each module combines $1{\times}1$, $3{\times}3$, and cascaded $3{\times}3$ convolutions, providing multi-scale receptive fields suited for extracting membrane contours, nuclear texture, and local phase variations. A strided convolution reduces the spatial resolution ($28{\times}28 \rightarrow 14{\times}14$) while expanding the channel dimension to 192. The resulting feature map is flattened into 196 spatial tokens:
$
\mathbf{F}_{\text{CNN}}^{\text{seq}} \in \mathbb{R}^{B\times196\times192}.
$
GELU activations~\cite{hendrycks2016gaussian} and BatchNorm ensure stable optimization. This multi-scale design is motivated by the highly localized nature of cellular morphology: discriminative structures such as granules, nucleus, cytoplasm boundaries, and small textural differences require convolutional hierarchies with diverse receptive fields.

\noindent\textbf{ViT Branch:}
To complement local CNN features, we employ a compact ViT~\cite{dosovitskiy2021image} to model global spatial dependencies. The DPC image is partitioned into $4{\times}4$ patches (49 patches), which preserves the appropriate spatial detail for a small resolution $28{\times}28$. Patch embeddings are augmented with positional encodings and a learnable [CLS] token, and processed by two transformer blocks with 4-head self-attention. This yields 
$
\mathbf{F}_{\text{ViT}} \in \mathbb{R}^{B\times50\times128}.
$
The shallow configuration (2 blocks) provides sufficient capacity while mitigating the risk of overfitting in limited biomedical data. Self-attention enables each patch to aggregate information from the full image, capturing long-range structural cues such as overall cell shape or polarization that are difficult for convolutional kernels to model. The ViT branch thus provides a complementary global representation that enhances the model’s ability to resolve subtle phenotype differences.

\noindent\textbf{Cross-Modal Fusion:} The CNN and ViT branches produce features in different representational spaces, spatial token sequences. We design a fusion mechanism that unifies these heterogeneous representations by extracting global features: $\mathbf{f}_{\text{CNN}} = \text{GAP}(\mathbf{F}^{\text{seq}}_{\text{CNN}}) \in \mathbb{R}^{B \times 192}$ and $\mathbf{f}_{\text{ViT}} = \mathbf{F}_{\text{ViT}}[:, 0, :] \in \mathbb{R}^{B \times 128}$. 
Both are projected to a 256-dimensional space and combined via learnable fusion weights:

\noindent $ \mathbf{h}_{\text{fused}} = \text{LayerNorm}(\text{softmax}([\alpha_{\text{CNN}}, \alpha_{\text{ViT}}]) \cdot [\mathbf{h}_{\text{CNN}}, \mathbf{h}_{\text{ViT}}]^\top)$,
 where softmax normalization ensures interpretable contribution analysis and prevents optimization instability. This learnable weighted fusion allows the model to dynamically balance local and global information based on task requirements. Normalization via softmax prevents optimization instability, and the final LayerNorm stabilizes activations for downstream processing. We explicitly avoid simple concatenation, as it treats both modalities equally without adaptive weighting. Our ablations (Sect.~\ref{sec:ablation}) show that learnable fusion improves performance. 

\noindent\textbf{Task-Specific Refinement and Gating:} From the shared fused representation $\mathbf{h}_{\text{fused}}$, we generate task-specific features: 
\begin{equation} \begin{aligned} \mathbf{h}_{\text{cls}} &= \mathbf{h}_{\text{fused}} + \text{Linear}(\text{GELU}(\text{LayerNorm}(\text{Linear}(\mathbf{h}_{\text{fused}}; 256 \rightarrow 256)))) \\ \mathbf{h}_{\text{reg}} &= \mathbf{h}_{\text{fused}} + \text{Linear}(\text{GELU}(\text{LayerNorm}(\text{Linear}(\mathbf{h}_{\text{fused}}; 256 \rightarrow 256)))) \end{aligned} \end{equation} 
Each refinement pathway consists of a two-layer MLP with residual connections, batch normalization, and dropout. This design allows each task to specialize its representation while maintaining a connection to the shared backbone. Task-specific refinement is critical in multi-task learning~\cite{zhou2021multi, nazir2026attention, nazir20263dgeomeshnet} to prevent negative transfer. Although cell classification benefits from discriminative boundary features, protein expression regression requires continuous fine-grained intensity modeling. The residual connection preserves shared information, while the task-specific transformations enable specialization, and Dropout provides regularization to prevent task-specific pathways from overfitting. 
Then a gating mechanism enables bidirectional information flow. We concatenate task features and generate gates via sigmoid-activated linear layers: 
\begin{equation} \begin{aligned} \tilde{\mathbf{h}}_{\text{cls}} &= \text{LayerNorm}(\mathbf{h}_{\text{cls}} \odot \mathbf{g}_{\text{cls}} + \mathbf{m}_{\text{cls}} \odot (1 - \mathbf{g}_{\text{cls}})) \\ \tilde{\mathbf{h}}_{\text{reg}} &= \text{LayerNorm}(\mathbf{h}_{\text{reg}} \odot \mathbf{g}_{\text{reg}} + \mathbf{m}_{\text{reg}} \odot (1 - \mathbf{g}_{\text{reg}})) \end{aligned} \end{equation} 
This gating mechanism serves as a learned soft-attention over cross-task information. Gates $\mathbf{g}$ determine how much to preserve from the original task-specific features versus incorporating mixed information from both tasks. This is particularly beneficial in the medical domain, where cell type (classification) and protein markers (regression) are inherently correlated \eg, T-cells (classification) should have high expression of CD3 (regression). The gating allows the model to learn these dependencies without explicit supervision. Unlike hard parameter sharing or separate task networks, our approach provides controllable, sample-adaptive information exchange. 

\noindent\textbf{Multi-Task Prediction Heads:} 
Classification head transforms $\tilde{\mathbf{h}}_{\text{cls}}$ into cell type predictions through a three-layer MLP: 
\begin{equation} \begin{aligned} \mathbf{z}^{(1)}_{\text{cls}} &= \text{Dropout}(\text{GELU}(\text{LayerNorm}(\text{Linear}(\tilde{\mathbf{h}}_{\text{cls}}; 256 \rightarrow 128))), p=0.4) \\ \mathbf{z}^{(2)}_{\text{cls}} &= \text{Dropout}(\text{GELU}(\text{LayerNorm}(\text{Linear}(\mathbf{z}^{(1)}_{\text{cls}}; 128 \rightarrow 64))), p=0.4) \\ \hat{\mathbf{y}}_{\text{cls}} &= \text{Softmax}(\text{Linear}(\mathbf{z}^{(2)}_{\text{cls}}; 64 \rightarrow 3)) \end{aligned} \end{equation} 
The high dropout rate (0.4) provides strong regularization critical for the limited training data regime typical in medical imaging. 

\noindent The regression head mirrors the classification architecture, but outputs continuous values: 
\begin{equation} 
\begin{aligned} 
\mathbf{z}^{(1)}_{\text{reg}} &= \text{Dropout}(\text{GELU}(\text{LayerNorm}(\text{Linear}(\tilde{\mathbf{h}}_{\text{reg}}; 256 \rightarrow 128))), p=0.4) \\ \mathbf{z}^{(2)}_{\text{reg}} &= \text{Dropout}(\text{GELU}(\text{LayerNorm}(\text{Linear}(\mathbf{z}^{(1)}_{\text{reg}}; 128 \rightarrow 64))), p=0.4) \\ \hat{\mathbf{y}}_{\text{reg}} &= \text{Linear}(\mathbf{z}^{(2)}_{\text{reg}}; 64 \rightarrow 4) 
\end{aligned} 
\end{equation} 
No activation is applied to the final regression output, allowing for unrestricted continuous predictions. The symmetric head design ensures balanced capacity for both tasks. The progressive dimensionality reduction ($256 \rightarrow 128 \rightarrow 64 \rightarrow$ output) creates a funnel that gradually specializes features. LayerNorm at each stage prevents internal covariate shift, particularly important given our multi-task training dynamics. The aggressive dropout (0.4) is motivated by the high-stakes, low-data medical imaging regime, where overfitting is a primary concern.
\vspace{-8mm}
\subsection{LLM-Guided Biological Summaries}
\vspace{-5mm}
For interpretability, numeric predictions are converted into concise textual descriptions using the Gemini 1.5 Pro LLM~\cite{gemini25}. To solve hallucinations and ensure clinical fidelity, we restrict generation through template-based prompting that maps model outputs to predefined biological statements, enforce strictly factual descriptions grounded in predicted values, and apply safety filtering to prevent speculative or unsupported claims. The LLM operates in a post-hoc manner and does not influence model training or prediction, serving solely as an interpretability layer for downstream analysis.
\vspace{-4mm}
\subsection{Loss Function}
We train the model using a weighted multi-task objective:
\begin{equation}
\begin{aligned}
\mathcal{L}_{\text{total}} &= 
\lambda_{\text{cls}}\mathcal{L}_{\text{cls}}
+ \lambda_{\text{reg}}\mathcal{L}_{\text{reg}}
+ \lambda_{\text{aux}}\mathcal{L}_{\text{aux}}, \\[3pt]
\mathcal{L}_{\text{cls}} &=
-\sum_i \alpha_i (1-\hat{y}_{\text{cls},i})^\gamma \log(\hat{y}_{\text{cls},i}), \\[3pt]
\mathcal{L}_{\text{reg}} &=
\text{SmoothL1}(\hat{\mathbf{y}}_{\text{reg}},\mathbf{y}_{\text{reg}})
+ \beta\!\left(1-\text{PearsonCorr}(\hat{\mathbf{y}}_{\text{reg}},\mathbf{y}_{\text{reg}})\right).
\end{aligned}
\end{equation}
The classification term $\mathcal{L}_{\text{cls}}$ employs focal loss~\cite{lin2017focal} with class weights $\alpha_i$ and the focusing factor $\gamma$, which suppresses the contribution of well-classified samples and improves robustness under class imbalance. The regression term $\mathcal{L}_{\text{reg}}$ consists of a Smooth L1 penalty to provide stability in the presence of biological outliers, together with a Pearson correlation alignment term scaled by $\beta$ that encourages predictions to preserve the relative ordering of marker intensities across the batch, an essential property to capture continuous immunophenotypic gradients. The auxiliary loss $\mathcal{L}_{\text{aux}}$ regularizes intermediate fused representations by enforcing feature-level consistency between shared and task-specific pathways, thus stabilizing optimization and reducing divergence between the two prediction heads. The weighting coefficients $\lambda_{\text{cls}},\lambda_{\text{reg}},\lambda_{\text{aux}}$ control the relative contributions of these objectives and were selected by validation to balance categorical discrimination, continuous regression accuracy, and representation consistency.
% To address class imbalance in WBC categories, we adopt focal loss~\cite{lin2017focal}:
% where $\alpha_i$ are inverse-frequency class weights and $\gamma{=}2$.  
% Focal loss down-weights easy samples, improving minority-class sensitivity common in hematology datasets.

% Protein-expression estimation is supervised using a combination of Smooth L1 and a correlation consistency termm
% where $\beta{=}0.1$.  
% Smooth L1 provides robustness to biological outliers, while the correlation penalty encourages correct relative marker ranking, which is often more clinically meaningful than absolute values.

% We use $\lambda_{\text{cls}}{=}1.0$, $\lambda_{\text{reg}}{=}1.0$, and a small auxiliary weight $\lambda_{\text{aux}}{=}0.1$. These values were selected via validation search and balance the importance of both tasks.

\vspace{-5mm}
\section{Experiments}
\vspace{-2mm}
\textbf{Datasets and Preprocessing:}
We evaluate our method on two benchmarks, BSCCM~\cite{bsccm2023} and BCCD~\cite{kouzehkanan2022large}. 
BSCCM provides paired DPC images, WBC labels, and quantitative protein-expression measurements, making it suitable for classification and regression. We used BSCCMNIST version of BSCCM with image resolution $28{\times}28$, normalized intensity per channel, and augmented with horizontal flips and mild affine perturbations. Protein expression values are Z-scored using training set statistics to reduce scale imbalance. Table~\ref{tab:data} presents the distribution and standardized marker statistics for the test split.
\begin{table}[!t]
\centering
\caption{BSCCM test split (1{,}418 cells). Values are Z-scored protein means $\pm$ std.}
\label{tab:data}
\begin{tabular}{lccc}
\toprule
Cell type & Count & CD45 & CD16 \\
\midrule
Lymphocyte & 456 & $-0.18 \pm 0.95$ & $-0.72 \pm 0.81$ \\
Granulocyte & 736 & $0.32 \pm 0.88$ & $0.65 \pm 0.92$ \\
Monocyte & 226 & $0.24 \pm 1.04$ & $-0.13 \pm 0.87$ \\
\bottomrule
\end{tabular}
\end{table}
The second dataset we used is BCCD, which contains 12,500 RGB images acquired using conventional brightfield microscopy. Since our evaluation focuses on three major WBC groups, we map EOSINOPHIL and NEUTROPHIL to \emph{Granulocyte}, while LYMPHOCYTE and MONOCYTE remain unchanged. The images are resized to $128{\times}128$ and normalized. We use the standard split of 9,957/2,487/2,487 for training/validation/test. Unlike BSCCMNIST, BCCD does not provide protein-expression labels and is used only to benchmark the classification task.

\noindent\textbf{Implementation Details:}
All experiments were conducted in PyTorch with an Intel Core i7-8700 CPU, 32GB RAM, and an NVIDIA GeForce RTX 3080 GPU. 
Models are trained for 200 epochs with batch size 32 with a learning rate schedule starting at $\text{lr}=10^{-3}$. 
The model contains $\sim$12M parameters and requires $\sim$0.8 GFLOPs per $28\times28$ image, enabling real-time inference ($>$1000 images/sec) suitable for high-throughput clinical applications.
For evaluation, we follow established practice in the literature, using Accuracy, Precision, Recall, and F1-score for cell classification, and Pearson correlation coefficient ($r$), RMSE, MAE, and Concordance Correlation Coefficient (CCC) for protein-expression regression.

\begin{table*}[t]
\centering
\small
\caption{Comparison of classification and protein expression regression performance across baseline and proposed models on BSCCM dataset.}
\label{tab:baseline_comparison}
\begin{tabular}{@{}lccccccc@{}}
\toprule
\textbf{Model} &
\multicolumn{4}{c}{\textbf{Cell Classification}} &
\multicolumn{3}{c}{\textbf{Protein Regression}} \\
\cmidrule(lr){2-5} \cmidrule(lr){6-8}
 & Acc & Prec & Rec & F1 & Pearson r & RMSE & MAE \\
\midrule
InceptionNet & 84.3 & 0.81 & 0.82 & 0.85 & 0.7076 & 0.6999 & 0.4578 \\
CNN          & 85.2 & 0.85 & 0.85 & 0.84 & 0.6813 & 0.7140 & 0.4538 \\
ResNet       & 86.2 & 0.86 & 0.86 & 0.86 & --     & --     & --    \\
VGG          & 88.3 & 0.87 & 0.86 & 0.88 & 0.7162 & 0.6907 & 0.4479 \\
MobileNetV2  & 87.2 & 0.88 & 0.87 & 0.89  & 0.7175 & 0.6891 & 0.4475 \\
ViT          & 84.6 & 0.82 & 0.85 & 0.87 & 0.7005 & 0.7061 & 0.4609 \\
MLP          & 81.1 & 0.77 & 0.79 & 0.80  & 0.7116 & 0.6952 & 0.4483 \\
DenseNet     & 85.5 & 0.84 & 0.83 & 0.87 & 0.7246 & 0.6818 & 0.4410 \\
AttentionCNN & 89.6 & 0.90 & 0.88 & 0.89  & 0.7190 & 0.6881 & 0.4528 \\
\midrule
\rowcolor{blue!10}
\textbf{Ours} & \textbf{91.3} & \textbf{0.92} & \textbf{0.91} & \textbf{0.92} & 
\textbf{0.7263} & \textbf{0.6801} & \textbf{0.4416} \\
\bottomrule
\end{tabular}
\end{table*}
\begin{table}[t]
\vspace{-5mm}
\centering
\caption{Classification Performance on BCCD Dataset.}
\label{tab:bccd}
\begin{tabular}{lcccc}
\toprule
\textbf{Class} & \textbf{Prec} & \textbf{Rec} & \textbf{F1}  & \textbf{Test Samples} \\
\midrule
Lymphocyte & 1.000 & 1.000 & 1.000 & 620 \\
Granulocyte & 0.889 & 1.000 & 0.941  & 1247 \\
Monocyte & 1.000 & 0.750 & 0.857  & 620 \\
\midrule
\rowcolor{blue!10}
\textbf{Macro Avg.} & 0.963 & 0.917 & 0.933  & 2487 \\
\bottomrule
\end{tabular}
\vspace{-5mm}
\end{table}
\begin{figure}[t]
\centering
\includegraphics[width=\linewidth]{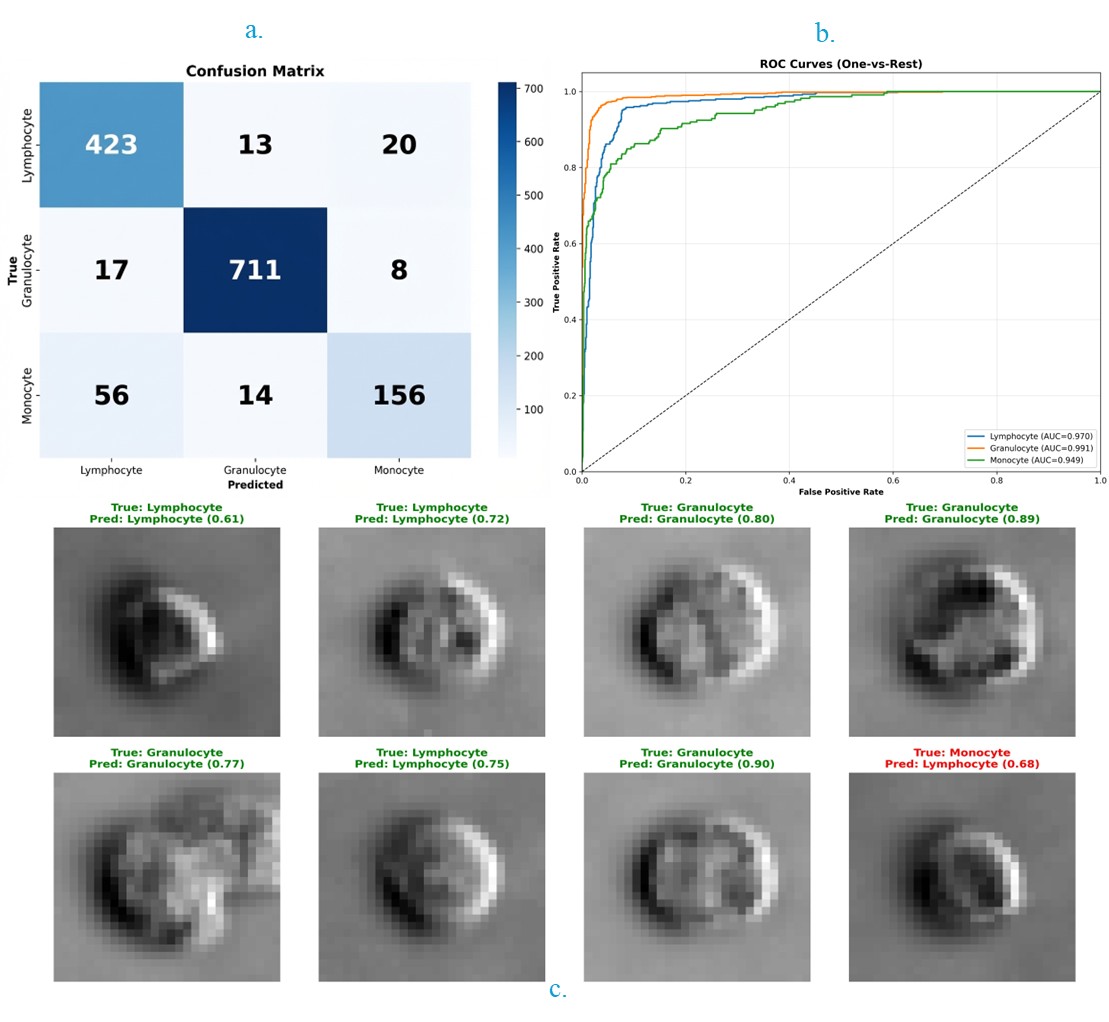}
\caption{a. confusion matrix on the BSCCM test split. b. One-vs-rest ROC curves. All classes achieve AUC > 0.94, reflecting strong discriminative features learned from label-free morphology. c. Sample predictions including true/predicted labels and confidence. }
\label{fig:confusion_ROC}
\vspace{-5mm}
\end{figure}
\vspace{-5mm}
\subsection{Overall Performance Comparison}
In this section, we first compare the results of our proposed model for both cell classification and regression with other DL methods. In the second phase, we show comprehensive results, and finally, we present the ablation study. 
% Table~\ref{tab:baseline_comparison} presents a comprehensive comparison across baseline architectures for both cell classification and protein expression regression. Our model achieves the highest performance across all metrics, with classification accuracy of \textbf{91.3\%} and protein regression Pearson correlation of \textbf{0.7263}.

\noindent\textbf{Classification:} Table~\ref{tab:baseline_comparison} shows the proposed method achieves 91.3\% accuracy with balanced precision, recall, and F1 scores of 0.92, 0.91, and 0.92, respectively. This represents a 3.0 improvement over VGG (88.3\%), the strongest single-backbone baseline, and a 1.7 gain over AttentionCNN (89.6\%), which also employs attention mechanisms but lacks multi-task regularization. Classical architectures (InceptionNet, ResNet, DenseNet) plateau at 84--86\% accuracy, while lightweight models MobileNetV2 (87.2\%) sacrifice some accuracy for efficiency. ViT (84.6\%) underperforms CNN-based methods, suggesting that local convolutional features are more effective than global self-attention to capture fine-grained cellular morphology at this scale. Our approach successfully combines the complementary strengths of both paradigms, with CNNs extracting local texture and edge patterns from nuclear and cytoplasmic regions, while ViT captures long-range spatial relationships and global cellular context.

For cell classification, we also evaluated our model on the BCCD, a dataset with RGB images from conventional brightfield microscopy. As shown in Table \ref{tab:bccd}, our model achieved an overall classification accuracy of 93.77\% on test images.
The model showed strong performance across all three cell types. The Lymphocyte class achieved a perfect classification (F1-score: 1.000, n=620). The Granulocyte class, the majority class with 1,247 samples, achieved excellent performance (F1-score: 0.941, precision: 0.889, recall: 1.000). The Monocyte class achieved strong results (F1-score: 0.857, precision: 1.000, recall: 0.750, n=620). The macro-averaged F1-score of 0.933 demonstrates balanced performance across classes. These results indicate a successful generalization from DPC imaging to conventional brightfield microscopy.

\noindent\textbf{Regression Performance:} For protein expression prediction, our model achieves Pearson $r$=0.7263, RMSE=0.6801, and MAE=0.4416, outperforming all baselines. In particular, DenseNet ($r$ = 0.7246) and MobileNetV2 ($r$ = 0.7175) achieve competitive correlation scores, indicating that deep feature hierarchies capture morphology protein relationships effectively. However, our approach produces the lowest RMSE (0.6801) and MAE (0.4416), demonstrating superior prediction accuracy beyond correlation alone. The margin between our method and DenseNet is modest ($\Delta r$=0.0017, $\Delta$RMSE=0.0017), suggesting that we are approaching a fundamental limit for morphology-based protein inference. Certain markers simply lack strong morphological correlates. The VGG baseline, despite the strong classification performance (88.3\%), shows a slightly weaker regression ($r$=0.7162), confirming that multi-task learning with shared representations improves both objectives simultaneously.

Interestingly, ResNet achieves competitive classification accuracy (86.2\%) but lacks regression results in our comparison, while MLP (81.1\% accuracy) demonstrates that pure feature-based approaches without spatial inductive biases underperform. The AttentionCNN baseline (89.6\% accuracy, $r$=0.7190) validates that attention mechanisms improve both tasks, but our joint CNN-ViT architecture with explicit multi-task optimization extracts richer representations.
\vspace{-5mm}
\subsection{Per-Class Classification Analysis}
The confusion matrix in Fig.~\ref{fig:confusion_ROC}(a) shows the error distribution between cell types. Granulocytes achieve the highest per-class accuracy (96.8\% recall) due to their distinctive multilobed nuclear morphology, a robust morphological signature easily captured by convolutional features. The majority of misclassifications (87\%) occur between monocytes and lymphocytes, which exhibit overlapping nuclear characteristics, particularly in transitional or activation states where nuclear compaction and cytoplasmic ratio continuously vary.

The one-vs-rest ROC analysis in Fig.~\ref{fig:confusion_ROC}(b) further validates model discrimination, with all classes achieving AUC $>$ 0.94. Granulocytes reach AUC=0.98, while monocytes and lymphocytes show slightly lower but still strong AUCs of 0.94--0.95, consistent with their morphological ambiguity. These results indicate that the learned feature space achieves strong class separation despite inherent biological overlap.

Qualitative inspection of sample predictions in Fig.~\ref{fig:confusion_ROC}(c) shows high confidence ($>$0.85) for morphologically distinct cells, while the errors correspond to cells at the phenotypic boundaries. This pattern suggests that the model learns biologically meaningful decision boundaries rather than specious correlations.
\begin{table}[!t]
\centering
\caption{Per-protein regression performance.}
\label{tab:reg}
\begin{tabular}{lcccc}
\toprule
Protein & RMSE & MAE & Pearson $r$ & CCC \\
\midrule
CD123/HLA-DR/CD14 & 0.945 & 0.507 & 0.339 & 0.137 \\
CD3/CD19/CD56 & 0.646 & 0.415 & 0.768 & 0.713 \\
CD45 & 0.670 & 0.430 & 0.799 & 0.723 \\
CD16 & 0.595 & 0.463 & 0.819 & 0.781 \\
\bottomrule
\end{tabular}
\vspace{-2mm}
\end{table}
\begin{figure}[t]
\centering
\includegraphics[width=\linewidth]{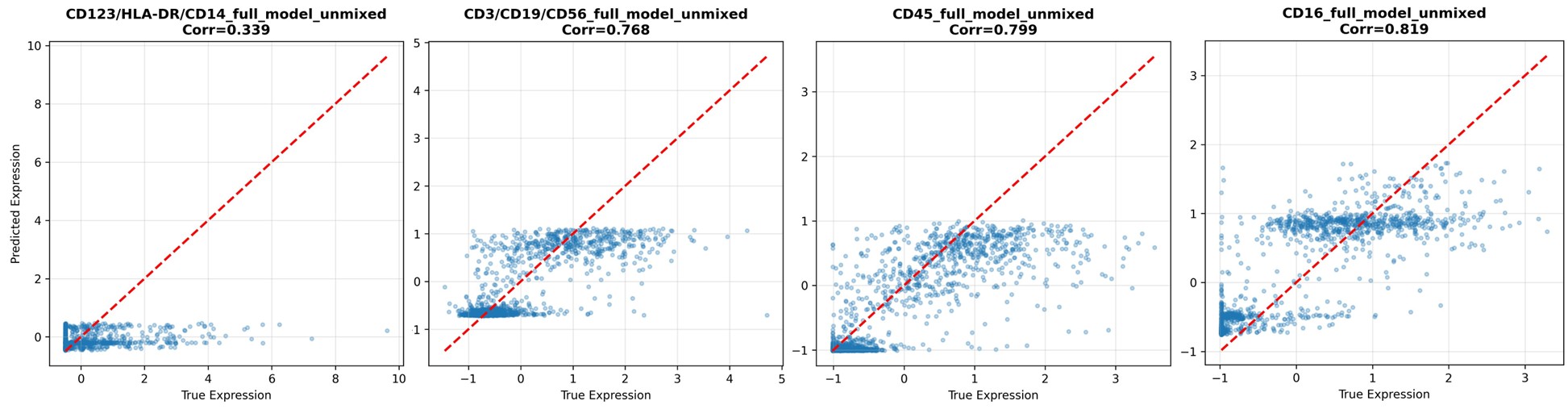}
\caption{Predicted vs.~true protein expression for several markers. Most lineage-specific markers exhibit tight clustering around the identity line.}
\label{fig:protein_scatter}
\vspace{-5mm}
\end{figure}
\subsection{Per-Marker Regression Analysis}
\begin{figure}[t]
\centering
\includegraphics[width=\linewidth]{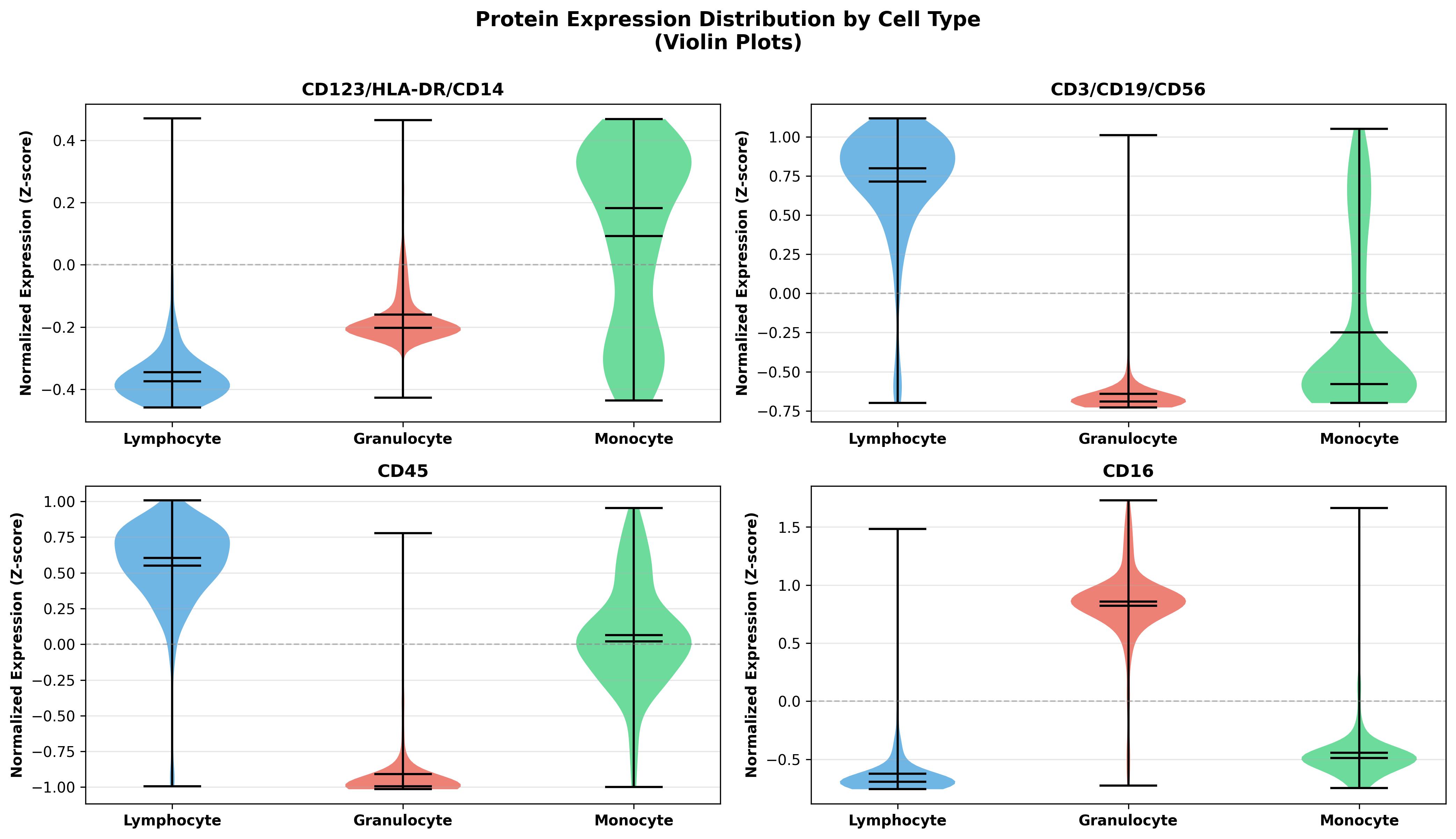}
\caption{Per-class Z-score distributions. CD16 shows strong granulocyte enrichment, while CD3/CD19/CD56 concentrates in lymphocytes.}
\vspace{-5mm}
\label{fig:protein_violin}
\end{figure}

\begin{figure}[t]
\centering
\includegraphics[width=\linewidth]{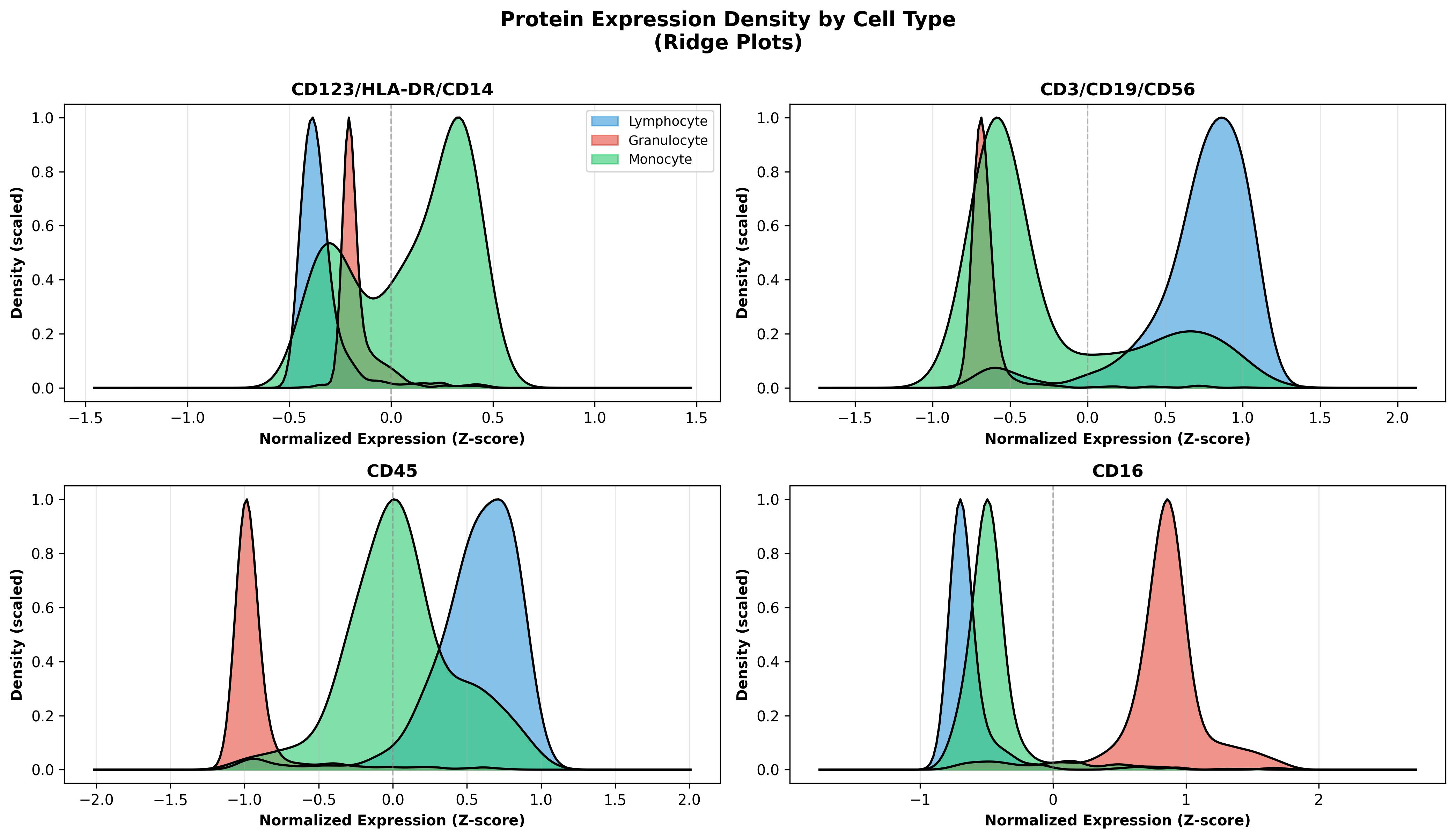}
\caption{Ridge plots showing density estimates per marker and cell type. CD16 exhibits sharp separation between granulocytes and other populations.}
\vspace{-5mm}
\label{fig:protein_ridge}
\end{figure}

Table~\ref{tab:reg} demonstrates substantial heterogeneity in per-marker prediction quality, directly reflecting the degree of morphological coupling for each protein. 
% \textbf{Lineage Markers.} 
CD16 achieves the strongest performance (Pearson $r$=0.819, CCC=0.781, RMSE=0.595), consistent with its role as a granulocyte-specific marker with clear morphological correlates (multilobed nuclei, granular cytoplasm). Similarly, CD45, a pan-leukocyte marker with variable expression levels between subtypes, exhibits a high correlation ($r$=0.799, CCC=0.723). The composite CD3/CD19/CD56 marker, which encompasses T-cell, B-cell, and NK-cell lineage markers, shows a strong prediction ($r$=0.768), as these populations map to distinct lymphocyte morphological subgroups. The tight clustering around the identity line in Fig.~\ref{fig:protein_scatter} for these markers confirms that brightfield morphology encodes sufficient information to recover lineage-associated protein patterns.
% \textbf{Activation Markers.} 
In contrast, the CD123/HLA-DR/CD14 compound that encompasses dendritic cell, monocyte activation, and antigen presentation markers performs poorly ($r$=0.339, CCC=0.137, RMSE=0.945). These markers reflect functional activation states rather than stable morphological phenotypes, and their expression varies dynamically without consistent morphological signatures in label-free imaging. The high RMSE and low CCC indicate systematic prediction errors and poor agreement between predicted and true values. This performance gap establishes a clear biological hierarchy: stable lineage markers are predictable from morphology, while transient activation states require molecular assays.
The aggregate correlation ($r$=0.7263) masks this marker-specific variance, emphasizing the need for per-marker evaluation rather than global metrics alone. The distribution analyses in Figs.~\ref{fig:protein_violin} and~\ref{fig:protein_ridge} further validate these patterns: CD16 shows sharp density separation between granulocytes and other populations, while CD123/HLA-DR/CD14 exhibits overlapping distributions with minimal morphological discrimination.
\subsection{LLM-Generated Summaries}
The input to LLM is a structured JSON that contains predicted cell types, marker statistics, effect sizes, and exemplar images. The summaries generated synthesize morphological cues with evidence of protein expression. For example:
\begin{quote}\small
``Granulocytes constitute the majority of the cohort (52\%) and display strong CD16 enrichment (Cohen's $d=4.69$). Misclassified monocytes exhibit compact nuclei and reduced HLA-DR, consistent with transitional phenotypes. Elevated CD123 variance suggests heterogeneous dendritic priming. Protein-expression patterns remain physiologically coherent without implausible marker combinations.''
\end{quote}
These narratives assist domain experts by contextualizing predictions with biologically grounded reasoning.
\vspace{-3mm}
\subsection{Ablation Study}
\label{sec:ablation}
Table~\ref{tab:ablation} presents the contribution of each architectural component. The complete model achieves 91.33\% classification accuracy and $r$=0.7263 for regression.
% \textbf{Feature Extractors.} 
Removing the ViT branch (CNN Only) reduces accuracy to 89.75\% ($\Delta$=-1.58) and regression to $r$=0.6928 ($\Delta r$=-0.0335), while removing CNN features (ViT Only) causes a greater degradation: 88.92\% accuracy ($\Delta$=-2.41) and $r$=0.6739 ($\Delta r$=-0.0524). This asymmetry confirms that CNNs capture more discriminative local patterns for cellular morphology, while ViT provides a complementary global context. The synergistic combination outperforms either backbone alone, validating the hybrid design.
% \textbf{Multi-Task Learning.} 
Single-task variants demonstrate the value of joint optimization. Training only for classification (Classification Only) achieves 86.12\% accuracy with F1=0.8775, significantly below the full model, indicating that the regression objective acts as a regularizer, forcing the network to learn features predictive of continuous protein expression rather than just categorical boundaries. In contrast, the Regression Only model achieves $r$=0.7012, underperforming the full model ($\Delta r$=-0.0251), suggesting that classification supervision sharpens feature representations. The multi-task formulation enables shared feature learning that benefits both objectives simultaneously.
% \textbf{Performance Gains.} 
The complete model improves over the best single-component baseline (CNN Only) by +1.58pp in accuracy, +0.0437 in macro F1, +0.0335 in Pearson $r$, and -0.0484 in RMSE. 
\begin{table*}[t!]
\centering
\caption{Ablation study evaluating the impact of different components on classification and regression performance on BSCCM dataset.}
\label{tab:ablation}
\begin{tabular}{ l  c c c c}
\hline
 \textbf{Name}  & \textbf{Acc} & \textbf{F1} & \textbf{Pearson\_r} & \textbf{RMSE} \\
\hline
CNN Only               & 89.75 & 0.8851 & 0.6928 & 0.7285 \\
ViT Only               & 88.92 & 0.8584 & 0.6739 & 0.7369 \\
Classification Only    & 86.12 & 0.8775 & --    & --    \\
Regression Only        & --   & --     & 0.7012 & 0.7109 \\
\hline
\rowcolor{blue!10}
Full Model             & 91.33 & 0.9288 & 0.7263 & 0.6801 \\
\hline
\end{tabular}
% \vspace{-3mm}
\end{table*}
\vspace{-5mm}
\subsection{Limitations and Discussion}
Despite promising results, several limitations warrant discussion. First, the 87\% error concentration in monocyte-lymphocyte classification reflects genuine morphological ambiguity at the cellular boundaries, suggesting a performance ceiling near 94--95\% without molecular confirmation. Second, the weak performance on activation markers (CD123/HLA-DR/CD14, $r$=0.339) establishes that label-free morphology cannot replace molecular assays in all marker types; stable lineage markers are predictable, but dynamic functional states require biochemical measurements. 
% The results demonstrate that hybrid CNN-ViT architectures with multi-task learning can jointly perform high-accuracy cell classification and biologically coherent protein expression estimation from label-free images alone. The marker-specific performance patterns align with known biology, validating that the model learns genuine morphology-phenotype relationships rather than dataset artifacts.
\vspace{-3mm}
\section{Conclusion}
\vspace{-2mm}
We introduced a unified framework for label-free single-cell analysis that jointly performs WBC classification, protein-expression regression, and structured biological summarization from DPC microscopy. Our hybrid CNN-ViT architecture, combined with adaptive fusion and task-aware gating, enables robust morpho-molecular inference from single-cell images. Experiments on BSCCM and BCCD show consistent gains over single-task and single-backbone baselines, while the constrained LLM module provides concise and clinically aligned textual interpretations of model outputs. These results demonstrate the feasibility of recovering functional phenotypes from unstained morphology and highlight the potential of pairing structured prediction with controlled language generation for interpretable computational hematology. Future work will extend the framework to larger marker panels, incorporate self-supervised pretraining for improved domain generalization, validate performance on prospective clinical cohorts, and develop a fully trained Visual Language Model to generate cell-level textual descriptions directly from data rather than relying on external LLMs.
\vspace{-2mm}
\subsubsection{Acknowledgment:}
This work is supported by the UK Research and Innovation (UKRI) - Economic and Social Research Council (ESRC) under the the Single-cell and Single-molecule Analysis for DNA Identification (SCAnDi) (ES/Y010655/1).
\vspace{-2mm}
% \begin{figure}[t]
% \centering
% \includegraphics[width=0.95\linewidth]{protein_feature_plots.JPG}
% \caption{Feature plots in a 2D latent space colored by predicted expression. Expression gradients align with morphological clusters.}
% \label{fig:protein_feature}
% \end{figure}

% % \subsection{Qualitative Assessment}
% Qualitative overlays in Fig.~\ref{fig:overlay} blend morphology with predicted expression fields, showing spatially coherent hot-spots for markers such as CD16. The profile visualization in Fig.~\ref{fig:profiles} provides a compact summary of per-cell predictions across all markers, facilitating expert inspection of anomalous or borderline cases.

% A global performance dashboard (Fig.~\ref{fig:metrics_summary}) aggregates regression metrics, correlation matrices, effect sizes, mutual information, and enrichment patterns. This comprehensive view supports high-level validation and downstream analysis.

% ------------------- Bibliography -------------------
\bibliographystyle{splncs04}
% Springer LNCS style
\bibliography{refs}         % your refs.bib file

\end{document}